\documentclass[]{jair}

\setcopyright{cc}
\copyrightyear{2025}
\acmDOI{10.1613/jair.1.18166}

\JAIRAE{Andrea Aler Tubella}
\JAIRTrack{Fairness and Bias in AI} 
\acmVolume{85}
\acmArticle{5}
\acmMonth{1}
\acmYear{2026}

\RequirePackage[
  datamodel=acmdatamodel,
  style=acmauthoryear,
  backend=biber,
  giveninits=true,
  uniquename=init
  ]{biblatex}
\RequirePackage{tcolorbox}
\tcbuselibrary{breakable}
\RequirePackage[normalem]{ulem}
\RequirePackage{multirow}
\RequirePackage{twemojis}
\RequirePackage{wasysym}

\begin{document}

\title[T-COL: Generating Counterfactual Explanations for General User Preferences]{T-COL: Generating Counterfactual Explanations for General User Preferences on Variable Machine Learning Systems}

\author{Ming Wang}
\orcid{0000-0001-8406-5677}
\email{sci.m.wang@gmail.com}
\affiliation{%
  \institution{Northeastern University}
  \city{Shenyang}
  \state{Liaoning}
  \country{China}
}

\author{Daling Wang}
\authornote{Corresponding Author.}
\orcid{0000-0003-1340-0778}
\email{wangdaling@cse.neu.edu.cn}
\affiliation{%
  \institution{Northeastern University}
  \city{Shenyang}
  \state{Liaoning}
  \country{China}
}

\author{Wenfang Wu}
\orcid{0000-0002-7215-563X}
\email{wenfang@stumail.neu.edu.cn}
\affiliation{%
  \institution{Northeastern University}
  \city{Shenyang}
  \state{Liaoning}
  \country{China}
}
\affiliation{
    \institution{University of G\"{o}ttingen}
    \city{G\"{o}ttingen}
    \state{Niedersachsen}
    \country{Germany}
}

\author{Shi Feng}
\orcid{0000-0002-2846-7652}
\email{fengshi@cse.neu.edu.cn}
\affiliation{%
  \institution{Northeastern University}
  \city{Shenyang}
  \state{Liaoning}
  \country{China}
}

\author{Yifei Zhang}
\orcid{0000-0003-0854-2966}
\email{zhangyifei@cse.neu.edu.cn}
\affiliation{%
  \institution{Northeastern University}
  \city{Shenyang}
  \state{Liaoning}
  \country{China}
}

\renewcommand{\shortauthors}{Wang, Wang, Wu, Feng \& Zhang}

\begin{abstract}
To address the interpretability challenge in machine learning (ML) systems, counterfactual explanations (CEs) have emerged as a promising solution. CEs are unique as they provide workable suggestions to users, instead of explaining why a certain outcome was predicted. The application of CEs encounters two main challenges: general user preferences and variable ML systems. On one hand, user preferences for specific values can vary depending on the task and scenario. On the other hand, the ML systems for verification may change while the CEs are performed. Thus, user preferences tend to be general rather than specific, and CEs need to be adaptable to variable ML models while maintaining robustness even as these models change. Facing these challenges, we propose general user preferences based on insights from psychology and behavioral science, and add the challenge of non-static ML systems as one preference. Moreover, we introduce a novel method, \uline{T}ree-based \uline{C}onditions \uline{O}ptional \uline{L}inks (T-COL) for generating CEs adaptable to general user preferences. Moreover, we employ T-COL to enhance the robustness of CEs with specific conditions, making CEs robust even when the ML models are replaced. To assess subjectivity preferences, we define LLM-based autonomous agents to simulate users and align them with real users. Experiments show that T-COL outperforms all baselines in adapting to general user preferences.
\end{abstract}



\received{22 January 2025}
\received[accepted]{22 November 2025}

\maketitle

\section{Introduction}
\label{sec:introduction}
Counterfactual explanations (CEs) offer a unique solution to address the interpretability limitations found in widely used and well-performing machine learning (ML) systems. Their implications extend to ML interpretability \cite{gunning2017explainable,arrieta2020explainable,gunning2019xai,molnar2020interpretable} and AI security \cite{sokol2019counterfactual,le2022improving,pmlr-v162-abid22a}, which are effective in increasing user trust in ML systems \cite{DELSER2024119898,METSCH2024104600}. First proposed by \citet{wachter2018counterfactual}, CEs aim to make minimal changes to the original data points, named as \textbf{query sample}, to achieve a desired classification outcome. It can also be considered as a new data sample with a desired category based on the derivation of the query sample.
For example, a loan decision ML system might reject a loan request from a user with a profile such as \{age: 24, education: bachelor, job: \textbf{service}, institution: private\}. Conventional methods might state that ``the rejection of your loan request is attributed to your service-oriented job'', whereas a CE would give a counterfactual like \{age: 24, education: bachelor, job: \textbf{profession}, institution: private\} and can be interpreted as ``having a professional job will result in loan approval''. This example illustrates that CEs reveal the specific differences in an instance that can lead to a desired outcome.
For its importance and uniqueness, many researchers have worked on the CE generation \cite{stepin2021survey,guidotti2022counterfactual,verma2020counterfactual,sokol2019counterfactual,artelt2019computation}. CEs are also used in various ML-based systems \cite{10.1145/3510457.3513081,albini2022counterfactual,yacoby2022if,piccione2022predicting,shang2022not,wellawatte2022model,smith2022individualized} and have garnered considerable attention within various ML communities \cite{tesic2022can,guidotti2022counterfactual,guidotti2019factual,filandrianos2022conceptual,dai2022counterfactual,pmlr-v151-pawelczyk22a,warren2022features,kuhl2022let}. 

However, the further application of CE still faces many challenges. The possibility of CEs is perceived as manipulative from a security view, as discussed in \cite{NEURIPS2021_009c434c}. The robustness of the CEs is analyzed in \cite{artelt2021evaluating} and \cite{virgolin2022robustness}.
\citet{verma2021counterfactual} summarize and list twelve key challenges for CEs in practical applications and industrial deployments. Inspired by it, we specifically address two of these challenges: challenge 3 of \textbf{non-static ML models} and challenge 7 of \textbf{capturing personal preferences}. On the one hand, users' preferences need to be captured by CE generation methods. Some researchers \cite{rasouli2022care} have attempted to incorporate user preferences into CEs by adding constraints on feature values. Alternatively, another approach \cite{9229232} built an interactive interface enabling users to select and set the range of feature values or tendencies for the features. Nevertheless, such specific user preferences for feature values can only reflect the user's tendency for a single task, making it difficult to adapt to complex tasks in real applications. For example, when applying for a loan, a person might tend to report a higher income, whereas when taxed, they might do the opposite. To address this issue, general preferences need to be captured. On the other hand, a common assumption in CE research is that ``the validation models for sample classification are fixed and do not change over time'', despite the frequent upgrades and changes observed in real-world ML systems.
The method in \cite{pmlr-v202-hamman23a} ensures the robustness of CE in variable ML systems by allowing arbitrary changes in the model parameter space while restricting predictive changes for points on the data manifold, and proposes \textit{stability} as a metric to measure this property. PROPLACE obtains an optimal solution that remains valid under all possible parameter variations, even in the presence of uncertainty in model parameters \cite{pmlr-v222-jiang24a}. However, these methods only consider changes in model parameters and do not fully account for model replacement. 
To research these two challenges in a unified way, we consider the robustness of CEs on variable ML systems as a general preference. In other words, it aspires to CEs with the highest possible success probability.

We investigate research in the fields of psychology and behavioral science \cite{benartzi2017smarter,chen2021market,bhatt2019attention}, and summarize the behavioral patterns of user preferences. With the help of psychologists, we categorize these patterns into 5 categories of general user preferences. They are abstract tendencies that are independent of specific tasks and reflect psychological traits of users.
To capture general user preferences, we propose a novel method called \uline{T}ree-based \uline{C}onditions \uline{O}ptional \uline{L}inks (T-COL) to generate CEs that capture general user preferences.
For a query sample, T-COL initially selects prototype cases based on user preference to construct a set of trees indicating the combinations of feature values derived from both the query sample and the prototype cases. Each tree contains a portion of the feature values. Then, T-COL selects the optimal combinations of local feature values based on an objective function designed for the user's preferences. The final CE is obtained by concatenating these local feature values and verification. Two components in T-COL are designed to capture user preferences: the prototype case screening and the local optimization objective. We denote a specific union of these two components as a \textbf{condition} of T-COL. We devised a separate condition for each preference, including one oriented to the issue of variable ML systems. Notably, these conditions are flexible and scalable. Simply by designing new conditions, T-COL can be adapted to new preferences.

To evaluate the adaptability of CEs generated by T-COL to general user preferences, we design a large number of Large Language Models (LLMs)-based agents to simulate user experiments with characterization prompt \cite{wang2024minstrelstructuralpromptgeneration}. By analyzing the behavioral consistency between LLM-based agents and real users in user research, we screened user-simulated agents (US-Agents). Guided by prompt engineering, US-Agents with different profiles are asked to choose the one that better matches their preferences among the CEs generated by different methods. To facilitate US-Agents to make choices, we evaluate the properties of CEs, such as \textit{proximity}, \textit{sparsity}, and \textit{validity}, as reference. Furthermore, we introduce \textit{centrality} and \textit{data fidelity} metrics to accommodate evaluations geared toward general user preferences.

In summary, our work has several primary contributions:

\begin{itemize}
    \item We propose general user preferences and an instance-based (IB) method, T-COL, with optional components to capture these preferences.
    \item We leverage user research findings to select and configure US-Agents. These agents then run simulated experiments to assess how well CEs adapt to user preferences.
    \item Extensive experiments show the strength of the CEs generated by T-COL in adaptation to variable ML systems and general user preferences.
\end{itemize}


\section{Related Work}
\label{sec:related_work}
In this section we introduce the generation methods and evaluation metrics in the research of CE.
\subsection{Counterfactual Explanation Generation Approaches}
The CE generation approaches can be divided into two main categories: Optimization (OPT) and Heuristic Search Strategy (HSS) \cite{guidotti2022counterfactual}. Among these, HSS can be further classified into instance-based (IB) and Decision Tree (DT)-based methods.

OPT approaches generate CEs by perturbing the original data points to cross the decision boundary and be classified as the desired class. There are model-agnostic optimization methods and model-specific computational methods designed for ML models \cite{artelt2019computation}. Some researchers model the generation of CEs as optimization problems with this idea and set different constraints for the objective \cite{10.1145/3527848}. For example, \citet{wachter2018counterfactual} introduced the concept of CE and constructed an optimization objective based on the distance between the counterfactual and the query sample and the prediction of the counterfactual made by the decision model, which was optimized by Adam \cite{kingma2014adam}. There are also researchers who have formalized most of the properties introduced in previous work on CE as different constraints on the optimization objective and solved for the CE using constrained optimization learning \cite{maragno2022counterfactual}. Optimization objectives can be solved using methods such as integer coding \cite{ustun2019actionable} and gradient descent \cite{mothilal2020explaining}. In addition, the solution of the CE problem has been formalized as an optimization problem for a non-monotonic submodular function in \cite{tsirtsis2020decisions}, which can be solved by randomization methods. \citet{guidotti_ensemble_2021} proposed an ensemble CE method that combines multiple weak counterfactual explainers to comprehensively consider all desired properties of CEs.

The methods used to solve optimization problems often operate on continuous values, leading to CEs that include unrealistic or understandable but not generally perceived features \cite{guidotti2022counterfactual}, such as half a master's degree getting a loan application or a person being 35.8 years old (around 35 years and 10 months, is an understandable feature value but not generally stated as such) having a high income. As a consequence, HSS methods have been developed to generate CEs based on features selected from the screened sample set or the feature space composed of all data samples. HSS methods involve solving for perturbations around query samples in the data space, where perturbations can be learned in the discrete data space by calculating diversity-enforcing losses \cite{Rodriguez_2021_ICCV} or finding the sample points most relevant to the target CE from the data space to construct counterfactual pairs, each counterfactual pair consists of a query sample and a selected target sample, and deriving the set of CEs from the counterfactual pairs by an iterative approach \cite{tran2021counterfactual}. \citet{silva2021teaching} generate CEs by sampling an uncertainty model to obtain a range of possible outcome states and selecting counterfactual states using a decision tree-like approach. \citet{keane2020good} propose an IB method to construct counterfactual pairs using good cases in the target category and use the different feature values in the counterfactual pairs as the corresponding feature values to compose CEs. Another method is diverse CEs by reusing the $ k $-nearest neighbor case pairs \cite{smyth2022few}, and \citet{goyal2019counterfactual} generated CEs of the bird figures by replacing features in the query samples at the corresponding location with features in the target category samples. In an extension of the IB approach, continuous features are transformed into categorical alternatives in \cite{warren2022features} to generate more effective CEs from the psychological perspective. For tree-structured classifiers, positive and negative paths are defined based on Boolean tests at internal nodes in \citet{10.1145/3097983.3098039}. By adjusting the feature vector, the classifier ensures that a decision tree predicting a negative outcome satisfies the Boolean conditions along the positive path.

Based on the characteristics of these two types of approaches, we adopt the IB method of HSS to design the CE generation method. Compared to the OPT-based approach, it can better ensure the feasibility and plausibility of CEs.

\subsection{Attributes and Evaluation Metrics}
How to evaluate the quality or explanatory effect of CEs has also received a great deal of attention from researchers. The properties and evaluation metrics of CEs, as well as the properties of interpreters, have been proposed.

\citet{verma2020counterfactual} formally and systematically presented the properties of CEs, containing metrics such as validity, actionability, and sparsity, providing metrics for research in related fields and standards for the application of CEs. Several properties on CE are expressed formally in \cite{maragno2022counterfactual} and defined in the form of formulas as constraints in the process of generating CEs. \citet{guidotti2022counterfactual} provided a very comprehensive summary of properties, presenting the properties associated with CE in terms of both CE and interpreters, respectively.

In addition, some researchers have also assessed CEs and interpreters from perspectives other than their properties. In order to improve the fairness of CEs, it is proposed in \cite{9660058} that the robustness of CEs must first be improved, and plausible CEs are defined to achieve this. \citet{laugel2019issues} also discuss the research work on CE from a robustness perspective. \citet{keane2021if} presented an analysis from a psychological and computational perspective, proposing requirements for distributionally faithful and instance-guided CEs. 

Building on these studies on properties and evaluation metrics, \citet{verma2021counterfactual} present 12 challenges for CE of future developments. However, there is currently little research on challenge 3 of \textbf{non-static ML systems} and challenge 7 of \textbf{capturing personal preferences}. For the two challenges, we propose to generate CEs under the conditions of general user preferences and variable ML models, and corresponding metrics to evaluate them.

\section{Problem Description}
\label{sec:definition}
Existing works \cite{9229232,rasouli2022care} only take the user's requirements for specific feature values into account, without a general consideration of the user's individual personality and preferences. However, in practical applications and deployments, users may not be able to express their requirements for specific features with certainty, but rather express their preference for a certain practice or a certain type of CEs. For example, users may be more likely to say ``I want an easy way to get approved for a loan'' rather than ``I want the type of work to be professional or managerial''.
To achieve this, we define general user preferences.

\subsection{General User Preferences}
\label{defination}
Drawing on research in psychology and behavioral science \cite{lee2024psychology,benartzi2017smarter,chen2021market,bhatt2019attention}, we propose the following general user preferences:

\begin{itemize}
    \item[\textit{A}] \textbf{Dedicated Preference}: I prefer to focus on a few things.
    \item[\textit{B}] \textbf{Minimalist Preference}: I wish it is easier to do.
    \item[\textit{C}] \textbf{Cautious Preference}: I want the solution with the highest success rate.
    \item[\textit{D}] \textbf{Admirer Preference}: I hope there are similar successful cases.
    \item[\textit{E}] \textbf{Collectivist Preference}: I hope the solution aligns with most successful cases.
\end{itemize}

General user preferences are not constraints on certain feature values and they are related to the user's personality and behavioral habits. As a result, these preferences do not usually change with tasks or scenarios.



\subsection{Problem Formulation}
To ensure \textit{feasibility}, we follow the IB approach for generating CEs. We selected feature values from the query sample and prototype cases to form new samples that meet the preference requirements, which serve as CEs. Let $ x $ denote a query sample, and $ \hat{x} $ denote an arbitrary combination of features. Then the CEs denoted as $ \Tilde{x} $ is the solution to (\ref{eq:1}).

\begin{equation}
\label{eq:1}
    \Tilde{x} = \arg \min_{\hat{x}} (|\operatorname{\textbf{M}}(\hat{x})-\Tilde{y}|+\operatorname{\textbf{d}}(\hat{x},x)) 
\end{equation}
where $ \Tilde{y}$ is the target class, $ \operatorname{\textbf{M}} $ denotes the ML model, and $ \operatorname{\textbf{M}}(\cdot) $ denotes the classification result of the model on the sample. In addition, $ \operatorname{\textbf{d}} $ denotes the distance between the counterfactual and the query sample, which can be calculated using any distance calculation function, such as Euclidean distance, Manhattan distance, Utility Space distance \cite{framling_contextual_2022}, etc. The first component represents the absolute value of the error between the classification result of the ML model $ \operatorname{\textbf{M}} $ for the current combination of features and the desired category. Reducing this value forces the solved CEs to reach the target category. The second component represents the distance between the current feature combination and the query sample, motivating the CEs generated to be closer to the query sample. The two constraints in (\ref{eq:1}) are the most essential ones, to which more external constraints can be added depending on the different requirements for the CEs.

Within IB CE explainers, the selection of feature values from different cases, which are selected samples in the target category, provides better assurance and control over the properties of the CEs and has therefore received more attention than the selection of feature values from all samples. In addition to the selection of feature values, the IB approach requires the screening of good cases as \citet{keane2020good} mentioned in the target category, to provide their feature values as the candidate set of feature values.

Based on the above analysis, we formalize the generation of CEs into the following three processes:
\begin{itemize}
    \item \textbf{Prototype case screening}: Select good samples from all samples in the target category based on the requirements for CEs as prototype cases, providing the candidate set of feature values to choose from when generating CEs.
    \item \textbf{Feature value selection}: Select a set or combination of sets of feature values from the screened cases and the query sample.
    \item \textbf{Feature value concatenating}: The selected feature values are concatenated in order while ensuring feature consistency, and CEs are obtained after validation.
\end{itemize}

Of the three processes listed above, \textbf{prototype case screening} and \textbf{feature value selection} have high correlations with the properties of CEs, which can be used to capture general user preferences with corresponding conditions.

\section{Methodology}
\label{sec:t_col}

T-COL captures general user preferences by selecting suitable conditions and generates CEs that match user preferences using processes such as prototype case screening and feature value selection. Additionally, to improve the efficiency of the explainer, we employ a partitioning strategy. T-COL first constructs \textbf{local greedy trees} using subsets of the divided local feature values to represent arbitrary local combinations of feature values. The feature values are derived from the query sample and prototype cases selected based on preferences. In addition, locally optimal objectives in conditions of feature values are selected on the \textbf{local greedy tree} according to general user preferences. 
Afterward, the feature value selection paths are obtained from \textbf{local greedy trees} and concatenated into a complete feature value selection path for selecting the feature values of CEs.

\subsection{General User Preference Analysis}

General user preferences are more like the personalities of the users rather than just requirements for a particular feature in a task. However, such general user preferences are abstract and can not simply be represented by feature values. They thus need to be linked to the properties of the CEs so that they can be captured in the generation of CEs. In other words, general user preferences can not be achieved directly by imposing constraints on the generation of CEs. Therefore, we first analyze the connection between general user preferences and the properties of CEs, as shown in (\ref{eq:mappings}).

\begin{equation}
    mappings = 
    \begin{cases}
        A : \{sparsity\},\\
        B : \{proximity\},\\
        C : \{actionability,~coherence,~validity\},\\
        D : \{data~manifold~closeness\},\\
        E : \{diversity,~data~manifold~closeness\}
    \end{cases}
    \label{eq:mappings}
\end{equation}

Dedicated preference \textit{A} requires as few different features as possible between CEs and the query sample so that the user can focus on changing a small number of feature values, requiring a higher degree of \textit{sparsity} in the generated CEs. Minimalist preference \textit{B} requires CEs to be easier to achieve, which intuitively means that the distance between CEs and the query sample should be as close as possible, placing a higher demand on the \textit{proximity} of CEs. Cautious preference \textit{C} indicates that the user wants to choose CEs that provide more secure solutions to achieve their needs, which requires a higher level of \textit{validity} of CEs. It is also necessary to ensure that the CEs have a higher degree of feasibility, validity, and consistency. Admirer preference \textit{D} indicates that the user wants similar success cases, which require CEs to be located in the feature space made up of existing data and to be as close as possible to the existing data, i.e., higher \textit{data manifold closeness}. Collectivist preference \textit{E} indicates that the user wishes to take the option chosen by the majority, i.e., CEs need to be representative and located in the feature space where the sample points are concentrated. Preference \textit{E} requires higher \textit{diversity} and \textit{data manifold closeness}.

\subsection{Overall Design of T-COL}
\begin{figure*}[h]
    \centering
    \includegraphics[width=\textwidth]{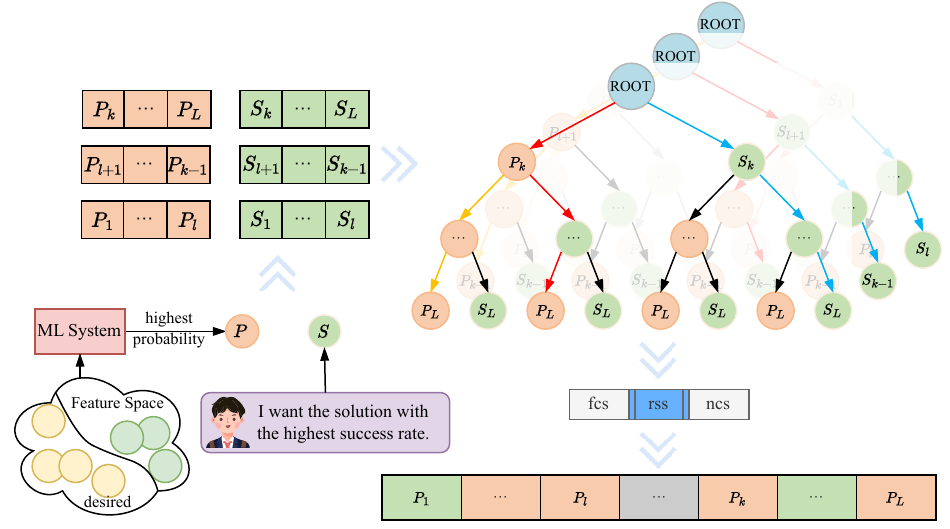}
    \caption{The whole process of T-COL, with the triangular arrow indicating the flow between the several processes. The user wants a more robust CE, and according to the previous introduction, T-COL first selects the target class sample for which the ML model gives the highest probability as the prototype case. After that, a group of local greedy trees is constructed using the prototype case and the query sample. For each tree, a combination of local feature values is selected using \textit{rss} and finally stitched into a CE.}
    \label{fig:T-COL}
\end{figure*}

To capture general user preferences, we introduce T-COL, including the prototype case screening and the feature value selection as optional conditions.
In order to improve the efficiency of T-COL, we use a partitioning strategy in the feature value selection process. We first screen prototype cases according to user preferences and construct \textit{local greedy trees} representing the local combination of feature values. Next, the local optimal objective is set according to the user's preference, and the optimal paths of local feature values are selected by using local greedy trees. Finally, all the paths selected by local greedy trees are concatenated in order. The complete path can guide the selection of feature values to compose counterfactuals from the query sample or prototype cases. The overall process is shown in Figure \ref{fig:T-COL}.

\begin{figure}
    \centering
    \includegraphics[width=0.6\textwidth]{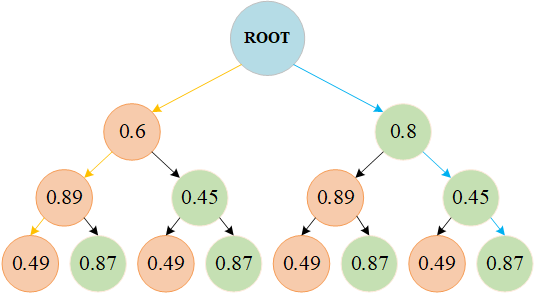}
    \caption{An example of a local greedy tree. The depth of the tree is equal to the number of elements in the local feature subset, and the nodes at each level store the feature values of the query sample and the prototype case, respectively. By traversing each path of the local greedy tree, several sets of local feature combinations can be obtained, and the optimal local feature combination can be selected according to the preset rules.}
    \label{fig:lgt}
\end{figure}

In Figure \ref{fig:T-COL}, the \textcolor[HTML]{01B0F0}{blue} dovetail arrows indicate the overall processes of T-COL. First, T-COL selects a prototype case (multiple prototype cases are available when users need diverse CEs) based on user preferences and divides its feature values with those of the query sample. After dividing the two samples, a number of subsets of local feature values are formed, with the \textcolor[HTML]{F7CBAC}{orange} and \textcolor[HTML]{C5E0B3}{green} nodes indicating the features of the prototype case and the query sample, respectively. The combination of the local feature values of the two samples is represented by constructing a local greedy tree (see Figure \ref{fig:lgt}). The \textcolor[HTML]{FF0000}{red} path is the optimal selection path from the local greedy tree.

If the user prefers more robust CEs, T-COL selects samples of the target category with the highest probability as prototype cases. In addition, `rss' (a local optimal objective, see Section \ref{sec:conditions}) will be used as a rule for feature value selection after the construction of local greedy trees. Ultimately, T-COL selects one or more combinations of feature values that match the user's preferences as CEs, from the set of feature values of the selected prototype cases and the query sample.

\subsection{Local Greedy Tree}
\label{sec:lgt}
A local greedy tree is a tree structure used to represent the combination of local feature values of the prototype case $ P $ and the query sample $ S $. The nodes of the tree consist of the feature values from a prototype case and the query sample.

For example, when the encoded feature vectors of the first three features of the two samples are $ \ddot{x}_l=[0.6,0.89,0.49] $ and $ x_l=[0.8,0.45,0.87] $ respectively, a local greedy tree can be constructed as shown in Figure \ref{fig:lgt}. The \textcolor[HTML]{F7CBAC}{orange} nodes indicate the encoded feature value of the prototype case $ P$, and the \textcolor[HTML]{C5E0B3}{green} nodes indicate the encoded feature value of the query sample $ S$. The \textcolor[HTML]{FFC000}{yellow} paths indicate the prototype case's local feature subsets, and the \textcolor[HTML]{00B0F0}{blue} paths indicate the local feature subsets of the query sample, while black paths denote other paths indicating arbitrary combinations of feature values.

A local greedy tree is a full binary tree, where the nodes of the tree can be classified into prototype and query nodes depending on their origin. Based on this, a locally greedy tree can be represented as $ LGT=Tree(V=(P\cup S), E) $. Then, $LGT$ can be used to select the local optimal feature value combination.

\subsection{Conditions for Capturing User Preferences}
\label{sec:conditions}
To construct a local greedy tree and use it to select local feature values, it is necessary to set up conditions, which contain the prototype case screening rules and the local optimal objectives.
The prototype case screening and feature value selection processes can be set up with different rules to suit general user preferences. On the basis of this, we further specify the concrete implementation of the two optional components corresponding to general user preferences for T-COL.

For dedicated preference \textit{A}, CEs with a smaller number of distinct feature values need to be generated. Therefore, when selecting the prototype cases we chose samples of the target category with fewer different feature values from the query sample as the prototype cases.
In addition, we define the feature value selection rule ``few-counterfactual score (\textit{fcs})'' as shown in (\ref{eq:fcs}).
\begin{equation}
    fcs = \operatorname{sigmoid}(\cos({\hat{x},\ddot{x}}))\times\displaystyle\sum^n_{i=1}\operatorname{equ}(|\hat{x}_i-x_i|)
    \label{eq:fcs}
\end{equation}
\begin{equation}
    \operatorname{equ}(x) = \begin{cases}
        1, & x=0 \\
        0, & x\neq 0
    \end{cases}
    \label{eq:equ}
\end{equation}
where $ n $ denotes the number of features in each sample, $ \Ddot{x} $ denotes a prototype case, and (\ref{eq:equ}) maps the difference in feature values to the difference in the number of feature values. The first component in (\ref{eq:fcs}) indicates the degree of resemblance between the current combination of feature values and the prototype case which means the probability that the current combination of feature values will be classified as the target class by the ML model, and the second part indicates the number of feature values that differ between the current combination of feature values and the query sample. \textit{fcs} combinations with the maximum or greater than a given threshold can be interpreted as alternate counterfactuals. We use a custom encoder that includes a scaler to encode the features, with categorical features being target-encoded \cite{pargent_regularized_2022}.

For minimalist user preference \textit{B}, which except the CEs to be as close as possible to the query samples, we choose the nearest target category sample to the query sample as the prototype case. The ``near-counterfactual score (\textit{ncs})'' is designed as the feature value selection rule, calculated as (\ref{eq:ncs}).
\begin{equation}
    ncs = \frac{\operatorname{sigmoid}(\cos(\hat{x},\Ddot{x}))}{\exp(\operatorname{d}(\hat{x}, x))}
    \label{eq:ncs}
\end{equation}
The above equation consists of two parts: the upper part represents the degree of similarity between the current combination of feature values and the prototype sample, and the lower part represents the distance between the current combination of feature values and the query sample. In addition, depending on the user's preference for CEs, we scale values using the exponential and sigmoid functions, respectively.

To accommodate cautious preference \textit{C} on the robustness of CEs on variable ML systems, the sample with the highest probability of being classified into the target category by the given ML model is chosen as the prototype case. In addition, we also follow the previous approach and give evaluation metrics for the combination of feature values in (\ref{eq:lrs}).
\begin{equation}
    rss = \frac{\exp(\cos(\hat{x},\Ddot{x}))}{\operatorname{sigmoid}(\operatorname{d}(\hat{x}, x))}
    \label{eq:lrs}
\end{equation}
To allow the generated CEs to be more securely classified as target categories, ``relative similarity score (\textit{rss})'' amplifies the importance of the similarity between the combination of feature values and the prototype case in (\ref{eq:lrs}).

In order to match the requirements of admirer preference \textit{D} that requires similar successful cases, the target category sample with the highest similarity to the query sample is selected as the prototype case, and \textit{rss} is used as the metric to evaluate the combination of feature values.

The largest number of samples can be found at the cluster centroid, so the sample near the cluster centroid is the optimal solution with respect to collectivist preference \textit{E}. Furthermore, \textit{rss} is equally applicable to the evaluation of the combination of feature values under such conditions.

\begin{table}[!t]
    \centering
    \caption{Local ``rss'' of each path in LGT}
    \begin{tabular}{c||ccc}
         \hline
         path         & similarity        & cost      & rss\\
         \hline
         \textless{}0, 0, 0\textgreater{} & \textbf{1}      & 0.6148 & 4.1882 \\
        \textless{}0, 0, 1\textgreater{} & 0.9682 & 0.4833 & 4.2572 \\
        \textless{}0, 1, 0\textgreater{} & 0.9466 & 0.4294 & 4.2542 \\
        \textless{}0, 1, 1\textgreater{} & 0.8757 & 0.2    & 4.3658 \\
        \textless{}1, 0, 0\textgreater{} & 0.9911 & 0.5814 & 4.2006 \\
        \textless{}1, 0, 1\textgreater{} & 0.9729 & 0.44   & 4.3495 \\
        \textless{}1, 1, 0\textgreater{} & 0.9128 & 0.38   & 4.1949 \\
        \textless{}1, 1, 1\textgreater{} & 0.8757 & \textbf{0}      & \textbf{4.8013} \\
        \hline
    \end{tabular}
    \label{tab:rss}
\end{table}

Using \textit{rss} as an example, the local feature value combination scores on each path of the local greedy tree in Figure \ref{fig:lgt} can be calculated, as shown in Table \ref{tab:rss}. By calculating the \textit{rss} of the combination of local feature values on each path of the local greedy tree, it can be concluded that the best CE path on these three feature value subsets is $ <1,1,1>$, which is the local feature value subset of the query sample $ S$.



\subsection{Utilizing T-COL to Address the Two Challenges}
According to the previous introduction, T-COL is an IB method with two optional components. It determines the selection rules for both prototype cases and local combinations of feature values based on general user preferences.
\subsubsection{General user preferences}
Capturing user preferences is the main purpose of designing T-COL and is mainly reflected in the two conditional optional components of T-COL, i.e., prototype cases screening and local optimal objective for feature values selection.

The first component controls the selection of prototype cases. In T-COL, the prototype cases are considered ideal cases about user preferences, and CEs are generated by directing the query sample to change toward such cases. The second component controls the selection of local feature value combinations. It can control how the query sample changes to the prototype case by controlling which feature values are selected.


In summary, T-COL generates CEs by controlling the query sample to change a certain degree in a certain direction. In which the direction and degree of change are determined by user preferences. For example, suppose a user wants to focus on a few things, T-COL will select samples with a few feature values different from those of the query sample as the ideal cases. In addition, T-COL selects as few different feature values from the prototype cases as possible during the change.

\subsubsection{Variable Machine Learning Systems}
Robust CEs on variable ML systems are designed to accommodate practical applications where the systems are frequently updated and changed. Instead of designing a model to deal with this issue, we treat it as a general user preference. As it expresses the user's expectation of CEs with a high success rate, we added Cautious preference \textit{C} to the general user preferences. Preference \textit{C} indicates the desire of users for more robust CEs, with the main issues being variable ML systems.

Based on T-COL, we set a condition for preference \textit{C}. We choose the target category samples with the highest classification confidence on the validation model, i.e., the highest classification probability of the validation model, as the prototype cases. In addition, when selecting a combination of local feature values, T-COL encourages the selection of more feature values from the prototype cases. By changing more in the direction of the highest confidence level, more robust CEs can be generated on variable ML systems.

\section{Experiments}
\label{sec:experiments}
To assess the adaptability of CEs to user preferences, we design US-Agents to simulate user experiments. The code, more experimental details, and reappearance methods are available at \url{https://github.com/sci-m-wang/T-COL}.
\subsection{Datasets}
Referring to the work on CE, we chose the five datasets: the Adult Income dataset \cite{kohavi1996scaling}, the German Credit dataset \cite{eggermont2004genetic}, the Titanic dataset \cite{titanic}, the Water Quality dataset \cite{water}, and the Phoneme dataset \cite{phoneme}. For more details about the datasets, please refer to Appendix \ref{appendix:dataset}.


\subsection{Metrics}
\label{sec:metrics}
To facilitate US-Agents' understanding of the task and their decision-making, we evaluate the \textit{proximity}, \textit{sparsity}, and \textit{validity} of CEs. Higher values indicate better \textit{validity}, while lower values are preferable for \textit{proximity} and \textit{sparsity}.

To better measure the adaptability of CEs to variable ML systems, we also propose a new property in addition to the above, called \textit{data fidelity}. It is defined as shown in (\ref{eq:df}).

\begin{equation}
    data \  fidelity = \frac{\displaystyle \sum^m_{i=1} (w_i \times p_i)}{\displaystyle \sum^m_{i=1} w_i}
    \label{eq:df}
\end{equation}

The aim of the proposed \textit{data fidelity} is to evaluate the validity of CEs when ML systems are changed, i.e., the ability of CEs to remain classified as the target class when the ML models change. Therefore, we used the classification results of CEs by third-party models (Arbitrary ML classification models) to assess the \textit{data fidelity} of CEs. In (\ref{eq:df}), $ m $ denotes the number of third-party models and $ w_i $ denotes the weight of the third-party models. The more discriminative it is of the original data, i.e. its classification accuracy as evaluated by the $ F1-score $, the greater its corresponding weight. Meanwhile, $ p_i $ denotes the classification accuracy of the third-party models for CEs expressed as $ F1-score $.

The $ F1-score $ is commonly used to assess the performance of a classifier, which we use to indicate the degree of endorsement of whether CEs belong to the target class. We show the \textit{data fidelity} of CEs generated by different methods in Table \ref{tab:pro}.

In addition to assessing general user preferences \textit{D} and \textit{E}, we propose the evaluation metric \textit{centrality}, expressed as the distance between the CE and the cluster centroid of the target category samples, in the form of (\ref{equ:cen}),

\begin{equation}
    centrality = \frac{1}{n}\times\displaystyle\sum^n_{i=1}\frac{\operatorname{d}(\Dot{x}_i,\Check{x})}{\operatorname{d}(\Dot{x}_i,\Tilde{x})}
    \label{equ:cen}
\end{equation}
where $ \Check{x}$ is the cluster centroid, $ \Dot{x}$ denotes the $ n$ sample points in the immediate vicinity of the cluster centroid, and $ \Tilde{x}$ denotes the CE. A higher \textit{centrality} of the CE indicates that the more similar it is to the majority of the target class samples, the better it matches the general user preferences.

\subsection{Baselines}
Three open-source methods for generating CEs are available in \textbf{DiCE} \cite{mothilal2020explaining} and can provide CEs of high quality on the available evaluation metrics. As a result, \textbf{DiCE} has been used as a baseline for most related studies. We follow the same approach, using \textbf{DiCE} as the baseline method. DiCE provides three methods to generate CEs, ``genetic'', ``random'' and ``kd-tree'', which we denote as ``DiCE-g'', ``DiCE-r'' and ``DiCE-k'', respectively. In addition, we compare the brute force method \cite{sokol2020fat-forensics} as a baseline method.


\begin{figure}[h]
    \centering
    \includegraphics[width=0.7\linewidth]{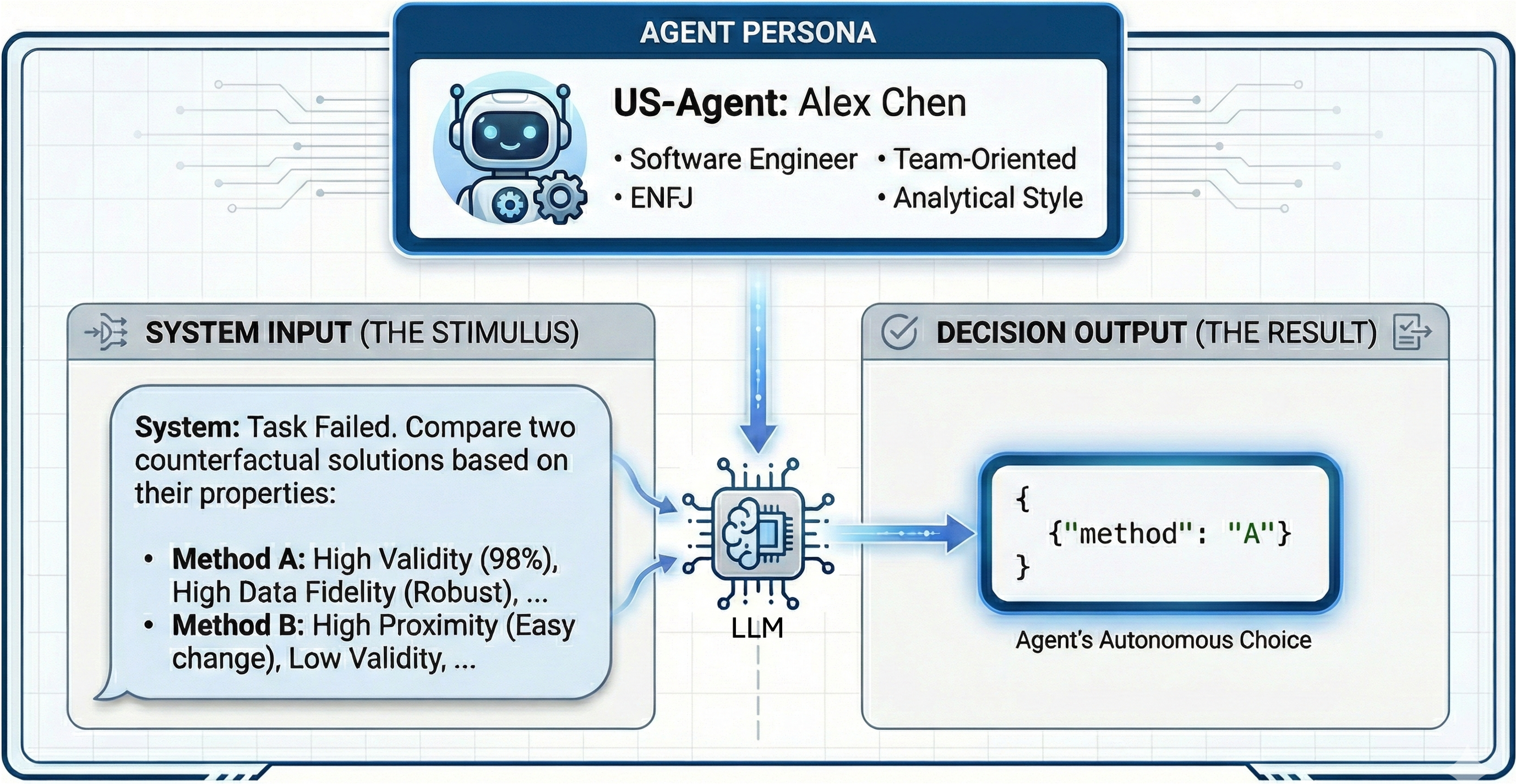}
    \caption{An example of a simulated user experiment for preference selection. To ensure the fairness of the experiment, T-COL and the baseline methods were replaced with method A or method B. Each US-Agent is asked to choose their preferred method, A or B, based on their preferences and the properties of CEs. To facilitate statistical analysis, each agent's response is restricted to JSON format.}
    \label{fig:simu-user_experiment}
\end{figure}

\subsection{Experimental Settings}
\label{sec:expsetting}
We randomly selected query samples, encoding the categorical features with Target Encoder \cite{micci2001preprocessing}. In the experiments, we generate five CEs for each query sample. For T-COL, the depth of local greedy trees was set to three.

\begin{table}[hb]
    \centering
    \caption{Validation weights of the third-party model}
    \begin{tabular}{c||ccccc}
    \hline
         Model & Adult Income & German Credit & Titanic & Water Quality & Phoneme\\
         \hline
KNN   & 0.73         & 0.66  &  0.93 & 0.62 & 0.84       \\
MLP   & 0.75         & 0.67  &  0.95 & 0.64 & 0.82       \\
SVM   & 0.74         & 0.69  &  0.93 & 0.62 & 0.8        \\
DT    & 0.69         & 0.65  &  0.94 & 0.58 & 0.83       \\
NB    & 0.74         & 0.7   &  0.93 & 0.52 & 0.75       \\
\hline
    \end{tabular}
    \label{tab:tpw}
\end{table}

A random forest model \cite{biau2016random} was chosen as the validation model for the CE to verify whether the generated CE was the target class. In addition, common ML models like Decision Tree \cite{safavian1991survey} and Naive Bayesian \cite{rish2001empirical} were selected as third-party models to evaluate the \textit{data fidelity} of the CEs. The weights of the third-party models were determined using a ten-fold cross-validation \cite{refaeilzadeh2009cross} approach. They are shown in Table \ref{tab:tpw}.

To evaluate the adaptability of CEs to general user preferences, we conduct simulated user experiments using LLM-based US-Agents. We first set profiles such as MBTI personality, birthday, occupation, etc. Additionally, we assign them examples of statements, symbol usage patterns, etc., referencing \cite{tsubota-kano-2024-text}. Then, we ask these agents with different settings to answer the questionnaire designed for real users. Finally, we compute the distributional consistency of the results of these LLM-based agents' responses to the questionnaire with the results collected in the real user research. For each real user, we screen the agents that have high consistency with them as US-Agents. After consistency screening, we select 23,987 US-Agents with InternLM \cite{cai2024internlm2}, Mistral \cite{mistral_ai_team_mistral-nemo-instruct-2407_nodate}, Qwen \cite{qwen2.5} and LLaMA \cite{grattafiori2024llama3herdmodels} as backbones. In other words, screened US-Agents made essentially the same choices on the same questionnaire as real users. An example of a simulated user experiment for preference selection is illustrated in Figure \ref{fig:simu-user_experiment}. An example of a complete US-Agent profile and prompts can be found in the Appendix \ref{appendix:us-agents}.

In the experiments, each agent is required to choose between T-COL and baseline methods according to the assigned preferences and the properties of CEs. T-COL is pitted against each baseline method, and the win rates are counted.

\subsection{Adaptability to General User Preferences}

To evaluate the adaptability of CEs generated by T-COL and baseline methods to general user preferences, we conduct simulated user experiments using LLM-based US-Agents. Each time a CE is selected, the corresponding generation method is credited with a win.
As references, we evaluate the properties of the CEs generated by the T-COL and baseline methods.
Since T-COL is an IB method and controls the source of feature values by screening prototype examples, it can effectively ensure the \textit{feasibility}, \textit{consistency}, and \textit{data manifold closeness} of CEs. Thus, we evaluate \textit{centrality}, \textit{data fidelity}, \textit{proximity}, \textit{sparsity}, and \textit{validity}. As described in Section \ref{sec:expsetting}, we generate CEs oriented to randomly selected query samples using different methods, and then compute their performance on these metrics.
The results are shown in Table \ref{tab:pro}.

\begin{table*}[h]
    \centering
    \caption{The contrasting properties of counterfactual explanations. BF denotes the brute force method. \textbf{Bolded values} indicate the best-performing results. `-1' indicates that all the CEs generated by this method are invalid.}
\resizebox{\textwidth}{!}{
\begin{tabular}{c|c|ccccccccc}
\hline
\multicolumn{1}{c}{\textbf{Datasets}} & \multicolumn{1}{c}{\textbf{Properties}} & \multicolumn{1}{c}{\textbf{BF}} & \multicolumn{1}{c}{\textbf{DiCE-g}} & \multicolumn{1}{c}{\textbf{DiCE-k}} & \multicolumn{1}{c}{\textbf{DiCE-r}} & \multicolumn{1}{c}{\textbf{T-COL-a}} & \multicolumn{1}{c}{\textbf{T-COL-b}} & \multicolumn{1}{c}{\textbf{T-COL-c}} & \multicolumn{1}{c}{\textbf{T-COL-d}} & \multicolumn{1}{c}{\textbf{T-COL-e}} \\ \hline
\multirow{5}{*}{Adult Income}         & centrality                              & 0.83                            & 0.91                                & 0.93                                & 0.99                                & 1                                    & 0.81                                 & 0.73                                 & \textbf{0.68}                        & \textbf{0.68}                        \\
                                      & data fidelity                           & 0.52                            & \textbf{0.74}                       & 0.67                                & 0.52                                & 0.36                                 & 0.45                                 & 0.62                                 & 0.68                                 & 0.71                                 \\
                                      & proximity                               & 1.2                             & 1.17                                & 1.22                                & 1.12                                & \textbf{0.96}                        & 1.09                                 & 1.09                                 & 1.07                                 & 1.07                                 \\
                                      & sparsity                                & 0.53                            & 0.45                                & 0.42                                & 0.45                                & \textbf{0.33}                        & 0.5                                  & 0.5                                  & 0.47                                 & 0.45                                 \\
                                      & validity                                & 0.57                            & 0.56                                & 0.56                                & 0.52                                & \textbf{1}                           & 0.91                                 & 0.94                                 & 0.98                                 & 0.96                                 \\ \hline
\multirow{5}{*}{German Credit}        & centrality                              & 0.88                            & 0.88                                & 0.96                                & 1.08                                & 1.02                                 & 0.89                                 & \textbf{0.73}                        & \textbf{0.73}                        & \textbf{0.73}                        \\
                                      & data fidelity                           & 0.61                            & 0.77                                & 0.73                                & 0.18                                & 0.6                                  & 0.71                                 & 0.78                                 & 0.78                                 & \textbf{0.79}                        \\
                                      & proximity                               & 1.3                             & 0.92                                & 0.89                                & 0.85                                & 0.84                                 & 0.66                                 & \textbf{0.65}                        & \textbf{0.65}                        & \textbf{0.65}                        \\
                                      & sparsity                                & 0.63                            & 0.43                                & 0.43                                & 0.5                                 & 0.35                                 & 0.26                                 & 0.25                                 & 0.26                                 & \textbf{0.23}                        \\
                                      & validity                                & \textbf{1}                      & 0.96                                & \textbf{1}                          & 0.36                                & \textbf{1}                           & 0.93                                 & 0.97                                 & 0.99                                 & 0.99                                 \\ \hline
\multirow{5}{*}{Phoneme}              & centrality                              & -1                              & 0.8                                 & 0.8                                 & 0.96                                & 1.27                                 & \textbf{0.75}                        & \textbf{0.75}                        & \textbf{0.75}                        & \textbf{0.75}                        \\
                                      & data fidelity                           & 0.12                            & 0.68                                & 0.69                                & 0.53                                & 1                                    & 0.72                                 & \textbf{0.76}                        & 0.74                                 & 0.74                                 \\
                                      & proximity                               & -1                              & 0.92                                & 0.92                                & 0.89                                & 1.3                                  & 0.54                                 & 0.54                                 & \textbf{0.5}                         & \textbf{0.5}                         \\
                                      & sparsity                                & -1                              & 1                                   & 1                                   & 0.88                                & 0.84                                 & \textbf{0.48}                        & \textbf{0.48}                        & \textbf{0.48}                        & \textbf{0.48}                        \\
                                      & validity                                & 0                               & \textbf{1}                          & \textbf{1}                          & 0.68                                & \textbf{1}                           & 0.96                                 & \textbf{1}                           & \textbf{1}                           & \textbf{1}                           \\ \hline
\multirow{5}{*}{Titanic}              & centrality                              & 0.88                            & 0.98                                & 0.87                                & 0.98                                & 0.86                                 & 0.65                                 & 0.65                                 & 0.65                                 & \textbf{0.62}                        \\
                                      & data fidelity                           & 0.49                            & 0.95                                & 0.98                                & 0.92                                & \textbf{1}                           & 0.99                                 & 0.99                                 & \textbf{1}                           & \textbf{1}                           \\
                                      & proximity                               & 1.23                            & 1.11                                & 1.17                                & 1.05                                & 1.32                                 & \textbf{1.03}                        & \textbf{1.03}                        & \textbf{1.03}                        & \textbf{1.03}                        \\
                                      & sparsity                                & 0.69                            & 0.49                                & 0.47                                & 0.47                                & 0.49                                 & \textbf{0.36}                        & \textbf{0.36}                        & \textbf{0.36}                        & \textbf{0.36}                        \\
                                      & validity                                & 0.54                            & \textbf{1}                          & 0.96                                & 0.96                                & \textbf{1}                           & \textbf{1}                           & \textbf{1}                           & \textbf{1}                           & \textbf{1}                           \\ \hline
\multirow{5}{*}{Water Quality}        & centrality                              & 0.88                            & \textbf{0.77}                       & 0.83                                & 0.85                                & 0.99                                 & 0.82                                 & 0.82                                 & 0.82                                 & 0.82                                 \\
                                      & data fidelity                           & 0.18                            & 0.47                                & 0.67                                & 0.75                                & \textbf{0.85}                        & 0.76                                 & 0.77                                 & 0.81                                 & 0.81                                 \\
                                      & proximity                               & 1.2                             & 0.97                                & 0.89                                & 0.99                                & 1.04                                 & 0.57                                 & 0.57                                 & 0.57                                 & \textbf{0.54}                        \\
                                      & sparsity                                & 1                               & 1                                   & 1                                   & 0.82                                & 0.73                                 & 0.47                                 & 0.47                                 & \textbf{0.42}                        & 0.44                                 \\
                                      & validity                                & 0.5                             & 0.84                                & \textbf{1}                          & 0.68                                & \textbf{1}                           & 0.97                                 & 0.97                                 & 0.97                                 & 0.97                                 \\ \hline
\end{tabular}
}
\label{tab:pro}
\end{table*}

Based on these results, we organize large-scale simulated user experiments to randomly match CEs generated by two methods, asking US-Agents to choose the one that better matches the specified user's preference. The results are shown in Table \ref{tab:pk}.

\begin{table}[]
    \centering
    \caption{Preference selection results from simulated user experiments, with values indicating the winning rate when T-COL confronts the baseline methods.}
    \resizebox{\textwidth}{!}{
    \begin{tabular}{c|cc|cc|cc|cc|cc}
\hline
\textbf{Preference}    & \multicolumn{2}{c|}{\textbf{A}} & \multicolumn{2}{c|}{\textbf{B}} & \multicolumn{2}{c|}{\textbf{C}} & \multicolumn{2}{c|}{\textbf{D}} & \multicolumn{2}{c}{\textbf{E}} \\ \hline
\textbf{Baseline}      & \textbf{DiCE}   & \textbf{BF}   & \textbf{DiCE}   & \textbf{BF}   & \textbf{DiCE}   & \textbf{BF}   & \textbf{DiCE}   & \textbf{BF}   & \textbf{DiCE}   & \textbf{BF}  \\ \hline
\textbf{Adult Income}  & 100.00          & 100.00        & 100.00          & 100.00        & 99.92           & 100.00        & 77.65           & 100.00        & 100.00          & 100.00       \\
\textbf{German Credit} & 94.08           & 100.00        & 100.00          & 100.00        & 53.63           & 99.83         & 50.21           & 79.65         & 64.97           & 99.75        \\
\textbf{Phoneme}       & 63.39           & 100.00        & 67.97           & 100.00        & 77.65           & 100.00        & 50.42           & 100.00        & 63.22           & 100.00       \\
\textbf{Titanic}       & 100.00          & 100.00        & 100.00          & 100.00        & 99.92           & 100.00        & 91.49           & 99.92         & 100.00          & 100.00       \\
\textbf{Water Quality} & 81.23           & 100.00        & 85.57           & 100.00        & 100.00          & 100.00        & 50.13           & 95.83         & 98.17           & 99.92        \\ \hline
\end{tabular}
    }
    \label{tab:pk}
\end{table}

The results show that the T-COL approach is more resilient to general user preferences. In addition, from the performance on \textit{data fidelity} metric and results of simulated user experiments, it is clear that T-COL can better address the challenges of variable ML systems compared to baseline methods.



\subsection{Efficiency Analysis}
The efficiency of generating CEs is also very important in applications. During our experiments, we found that the efficiency of DiCE and T-COL in generating CEs differed significantly. Therefore, we analyzed the time complexity of T-COL and DiCE and recorded the actual execution time of the experiment to generate five CEs for each query sample.

\subsubsection{Time Complexity Analysis}
Define $ m$ as the number of CEs generated for each sample, the number of features as $ n$, $ k$ as the number of samples generated during evaluation, and $ T$ as the number of iterations required to optimize the loss function. To facilitate the analysis of the time complexity of T-COL, we define $ d$ as the depth of the local greedy tree.

The main time-consuming steps in DiCE's process of generating CEs are:
\begin{itemize}
    \item Optimizing the loss function for counterfactual generation.
    \item Approximating the counterfactual to the decision boundary of the ML model.
\end{itemize}

The execution time of the first step is related to the number of CEs to be generated and the number of iterations. For each CE to be generated, a target loss value can be obtained based on the loss function. DiCE optimizes this loss value by means of an iterative approach, where each iteration requires the calculation of the distance between the query sample and the generated counterfactual. The time complexity of this part is $ \textsc{O}(T*mn)$. In the second step, a large number of samples need to be generated around the query sample at different distances. In addition, the distance from the query sample to all counterfactuals needs to be computed. The time complexity of this part is $ \textsc{O}(kmn)$. In summary, the time complexity of DiCE is $\textsc{O}((T+k)*mn)$.

\begin{table*}[]
    \centering
    \caption{Runtime (seconds) for generating CEs}
    \resizebox{\textwidth}{!}{
    \begin{tabular}{c|ccccccccl}
    \hline
    Datasets      & T-COL-a & T-COL-b & T-COL-c       & T-COL-d & T-COL-e       & DiCE-r  & DiCE-g & DiCE-k   &BF\\ \hline
    Adult Income  & 0.3     & 0.29    & \textbf{0.05} & 0.27    & 0.06          & 0.61    & 16.19  & 1.05     &4.45\\
    German Credit & 0.53    & 0.52    & \textbf{0.1}  & 0.47    & \textbf{0.1}  & 0.85    & 2.36   & 19.54    &871.45\\
    Titanic       & 0.30    & 0.29    & \textbf{0.05} & 0.27    & 0.06          & 0.85    & 2.61   & 9.36     &10.25\\
    Water Quality & 0.17    & 0.17    & \textbf{0.03} & 0.15    & \textbf{0.03} & 3706.03 & 0.31   & 1373.27  &2021.22\\
    Phoneme       & 0.13    & 0.13    & \textbf{0.02} & 0.12    & \textbf{0.02} & 605.43  & 0.39   & 600.4    &0.22\\ \hline
    \end{tabular}
    }
    \label{tab:run_time}
\end{table*}

The main time-consuming processes of T-COL are the second and third steps, while the first step of feature partitioning takes very little time and can be ignored. For the generation of each counterfactual explanation, the second step requires the computation of the values corresponding to the features on all paths of each local greedy tree, with a time complexity related to the depth of the tree and the number of features in the samples as $ \textsc{O}(m*\frac{n}{d}*2^{(d-1)})$. The third step requires traversing all the generated counterfactual paths and performing feature selection with a time complexity of $ \textsc{O}(mn)$. Note that according to the setup of this paper, $d$ is a constant between three and nine, which gives $ 1 < \frac{1}{d}*2^{(d-1)} < 29$, and the time complexity of T-COL can be abbreviated as $\textsc{O}(mn)$.

\subsubsection{Runtime Comparison}
Based on the previous analysis, the time complexity of T-COL is much smaller than that of DiCE since it does not require a lot of optimization computations. In this section, we list the actual running time of the two methods in our experiments in Table \ref{tab:run_time}, for different datasets.

As shown in Table \ref{tab:run_time}, T-COL generates CEs much more efficiently than DiCE, allowing for real-time response, which is very important in applications. With a local greedy tree depth of 3, T-COL takes less than a second to generate all five CEs for each query sample while ``DiCE-r'' even takes more than 3,700 seconds to generate five CEs for a sample query in Water Quality.

In our experiments, we found that the generation time difference of T-COL shows a strong correlation with the number of sample features for a fixed local greedy tree depth, which is consistent with the results of our time complexity analysis. In addition, we find that the time required by different optimization methods for DiCE varies significantly in the face of different data types. The ``random'' method is faster to generate when there are many categorical features, while the ``genetic'' method is faster to generate when there are many numerical features.

\section{Conclusion}

\label{sec:conclusion}
In this paper, we define general user preferences and propose T-COL, an instance-based method for generating CEs to capture these preferences.
We set different conditions for prototype screening and the local optimal objective of T-COL to generate CEs that can better adapt to different general user preferences. Furthermore, we investigate generating more robust CEs for variable ML systems. To solve this problem, we incorporate a user preference and establish a condition to guide T-COL. Moreover, we evaluate the adaptability of CE generation methods to preferences by conducting simulated user experiments with LLM-based US-Agents that are consistent with the performance of real users. Our experiments on five benchmark datasets demonstrate that T-COL better adapts to different user preferences and effectively handles variable ML systems.

\section*{Limitations and Future Work}
While T-COL demonstrates effectiveness in generating user-adaptive counterfactuals, our approach has two inherent limitations that warrant further investigation:

\paragraph{Empirical Basis for Function Combinations} The selection of specific functions is primarily driven by our empirical effectiveness in capturing distinct aspects of user preferences, rather than theoretical uniqueness. Although these choices are widely used in the field of explainable AI, alternative functions (e.g., logarithmic scaling for cost sensitivity, hyperbolic tangent for bounded preferences) may prove equally or more suitable in certain domains. This design flexibility—while enabling adaptability—introduces subjectivity in function selection. Future work should establish rigorous criteria for function selection and explore domain-specific formulations.

\paragraph{Discrepancy Between Counterfactual Distance and Practical Actionability} In the research, to simplify the problem, we treat the counterfactual distance as equivalent to the difficulty of achieving the objective, which does not inherently reflect real-world implementation difficulty. This decoupling of geometric distance from true actionability represents a fundamental challenge in counterfactual explanation research. T-COL partially mitigates this through the flexible design of conditions, but deeper integration of domain knowledge is needed to bridge this gap.

In future work, we plan to expand T-COL by adding general user preferences to accommodate complex real-world scenarios. We would like to add further constraints on T-COL and enrich its structure to guarantee CEs' properties better and address emerging challenges. In addition, we will further customize the condition settings under each user preference. Furthermore, we note that the CEs generated by T-COL do not perform consistently with the conditions in terms of correspondence properties. Thus, we will further explore the relationship between general user preferences and the properties of CEs.

\begin{acks}
The work was supported by the National Natural Science Foundation of China (62172086,  62272092).
\end{acks}

\printbibliography

\appendix

\section{The User Research}
\label{appendix:user-research}
In this section, we outline more specific details regarding user research.
\subsection{Questionnaire}
\label{appendix:questionnaire}
To verify the validity and completeness of our proposal and to analyze the tendency of different real users towards these preferences, we designed the following questionnaire and organized a user survey.

\begin{tcolorbox}[colback=yellow!10!white, colframe=brown!70!olive, title=Problem Solving Preference Research, breakable, fontupper=\scriptsize]
\textbf{1. Your Gender:}
\begin{itemize}
    \item[\Circle] Male
    \item[\Circle] Female
\end{itemize}
\textbf{2. Your Age:}
\begin{itemize}
    \item[\Circle] Below 18
    \item[\Circle] 18 to 25
    \item[\Circle] 26 to 30
    \item[\Circle] 31 to 40
    \item[\Circle] 41 to 50
    \item[\Circle] Over 50
\end{itemize}
\textbf{3. Your Country:}
\vspace{0.5\baselineskip}

\hspace{0.5cm}\textless Select a country\textgreater
\vspace{0.5\baselineskip}

\textbf{4. Your Sector:}
\vspace{0.5\baselineskip}

\textless Select a sector from the list of ``Manufacturing'', ``Construction'', ``Education/Training'', etc. \textgreater
\vspace{\baselineskip}

\textbf{CAUTION}: \textit{Choose up to two for all of the following multiple-
choice questions, and one is allowed.}
\vspace{\baselineskip}

\textbf{5. If you failed at something, and there was a magic machine that could tell you the solution. Which solution would you favour?}
\begin{itemize}
    \item[\Square] Wish to work on one aspect to make it happen, don't change it in many ways.
    \item[\Square] Wish to make it as easy as possible, can be multi-faceted, but preferably change only a little in each area.
    \item[\Square] A solution with the highest possible success rate.
    \item[\Square] Solutions that have success stories.
    \item[\Square] What most people do.
    \item[\Square] Others.~\textless Additional content required\textgreater
\end{itemize}

\textbf{6. Suppose you are a drug researcher and you have the following reference scenarios after a failed development.}
\begin{itemize}
    \item[\Square] Only a few structural differences from the programme you designed.
    \item[\Square] Less differences between structures compared to yours.
    \item[\Square] Structure with the highest success rate.
    \item[\Square] Structure with success stories.
    \item[\Square] Structure adopted by most people.
    \item[\Square] Others.~\textless Additional content required\textgreater
\end{itemize}

\textbf{7. If an application for a loan is rejected, there are several successful cases available to you. Which one would you choose?}
\begin{itemize}
    \item[\Square] Different from your situation in only a few ways.
    \item[\Square] Very much like your situation.
    \item[\Square] Option most likely to be successful.
    \item[\Square] Option that has been applied successfully before.
    \item[\Square] Most people's situation.
    \item[\Square] Others.~\textless Additional content required\textgreater
\end{itemize}

\textbf{8. Suppose you are a shopper and you can't always pick the items that your customers like, and there are a few exemplary shoppers to learn from, who would you learn from?}
\begin{itemize}
    \item[\Square] Different from you in a few ways.
    \item[\Square] Very much like you.
    \item[\Square] A sales champion.
    \item[\Square] The shopper who successfully sells the product.
    \item[\Square] The average, popular shopper.
    \item[\Square] Others.~\textless Additional content required\textgreater
\end{itemize}

\textbf{9. Suppose you were a teacher with a class of underachieving students and there were other teachers you could ask for advice, who would you ask?}
\begin{itemize}
    \item[\Square] Different from you in a few ways.
    \item[\Square] Very much like you.
    \item[\Square] A top model teacher.
    \item[\Square] Teachers with better teaching results.
    \item[\Square] Ordinary teachers.
    \item[\Square] Others.~\textless Additional content required\textgreater
\end{itemize}

\textbf{10. Suppose you are a farmer, and the locust plague always fails in prevention. There are some programs, which one would you learn?}
\begin{itemize}
    \item[\Square] The farm is different from you in a few ways
    \item[\Square] The farm situation is very much like yours
    \item[\Square] Farms of famous farmers
    \item[\Square] Farms that have effectively prevented locust plagues
    \item[\Square] Most farms
    \item[\Square] Others.~\textless Additional content required\textgreater
\end{itemize}

\end{tcolorbox}
The first four questions in the questionnaire dealt with the user's personal information, such as age, gender, nationality, and the industry in which he or she is engaged. The latter questions correlate well with the general user preferences we predefined. In addition to a generic question, we set up scenarios in five domains - medicine, finance, sales, education, and agriculture - with options associated with general user preferences.

\subsection{User Profile}

Considering user privacy and ethical issues, we do not show specific user samples directly. As an alternative, we show the proportions of different user characteristics.

\subsubsection{Gender}
\begin{table}[h]
\centering
\caption{Gender distribution in user research.}
\begin{tabular}{cc}
\hline
Gender & Proportion (\%) \\ \hline
Male   & 80.6       \\
Female & 19.4       \\ \hline
\end{tabular}
\label{tab:user_profile-gender}
\end{table}


The gender distribution in the real user research is shown in Table \ref{tab:user_profile-gender}. The number of women in the user study was relatively small due to the demographic composition of the audience groups promoted. However, there is no order of magnitude significant deviation between the minority and the majority, and it still expresses to some extent the tendency of the users towards these preferences. In future work, we will also consider further expanding the groups included in the user research and balancing the gender distribution.

\subsubsection{Age}
\label{sec:age}

As shown in Table \ref{tab:user_preference-age}, the user research we organized covered the entire age group. Among them, the research subjects are mainly concentrated between the ages of 18 and 40.

\begin{table}[]
\centering
\caption{Age distribution in user research.}
\label{tab:user_preference-age}
\begin{tabular}{c|cccccc}
\hline
Age             & Below 18 & 18$\sim$25 & 26$\sim$30 & 31$\sim$40 & 41$\sim$50 & Over 50 \\ \hline
Proportion (\%) & 1.49     & 37.31      & 32.84      & 19.40      & 4.48       & 4.48    \\ \hline
\end{tabular}
\end{table}

\subsubsection{Region}

\begin{figure}[h]
    \centering
    \includegraphics[width=0.5\linewidth]{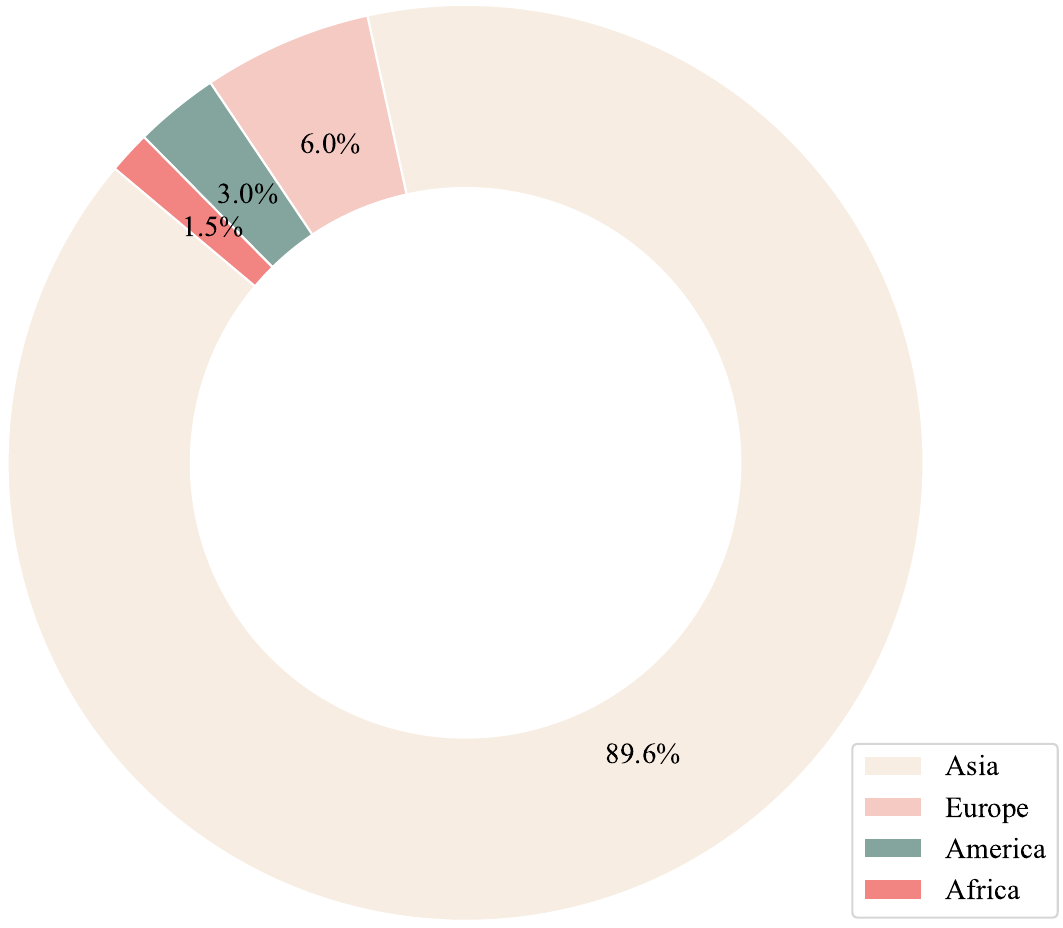}
    \vspace{-0.5em}
    \caption{The regional distribution of user research. The regions are grouped by continent.}
    \label{fig:user_research-region}
\end{figure}

In the user research, most users are from Asia, such as China, Singapore, Pakistan, etc. In addition, there are also some users from the United States, France, Germany, etc. The specific distribution is shown in Figure \ref{fig:user_research-region}.

\subsubsection{Sector}

\begin{figure}
    \centering
    \includegraphics[width=0.7\linewidth]{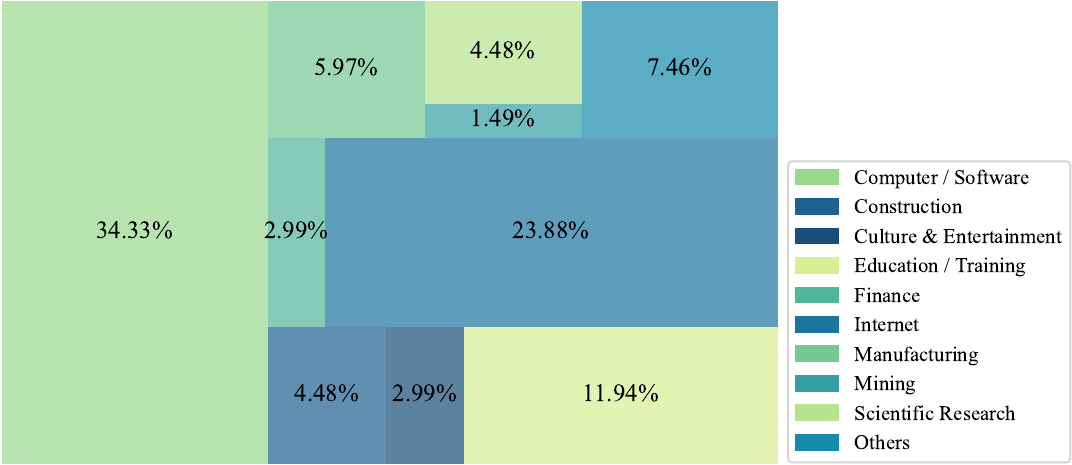}
    \caption{The industry distribution in user research. Others include industries with few participating users in the research, such as \textit{Healthcare or Social Security}, \textit{Logistics and Transport}, \textit{Professional Services} (e.g., legal or consultancy services), \textit{Wholesale and Retail}.}
    \label{fig:user_research-sector}
\end{figure}

As shown in Figure \ref{fig:user_research-sector}, the user research covered practitioners from a wide range of industries, with the computer and Internet industries dominating.

Above all, the composition of the respondents for the user research we organized is very diverse, and it can be thought that the results of the research are universal.

\subsection{User Preference Research}
\begin{figure}[h]
    \centering
    \includegraphics[width=0.7\textwidth]{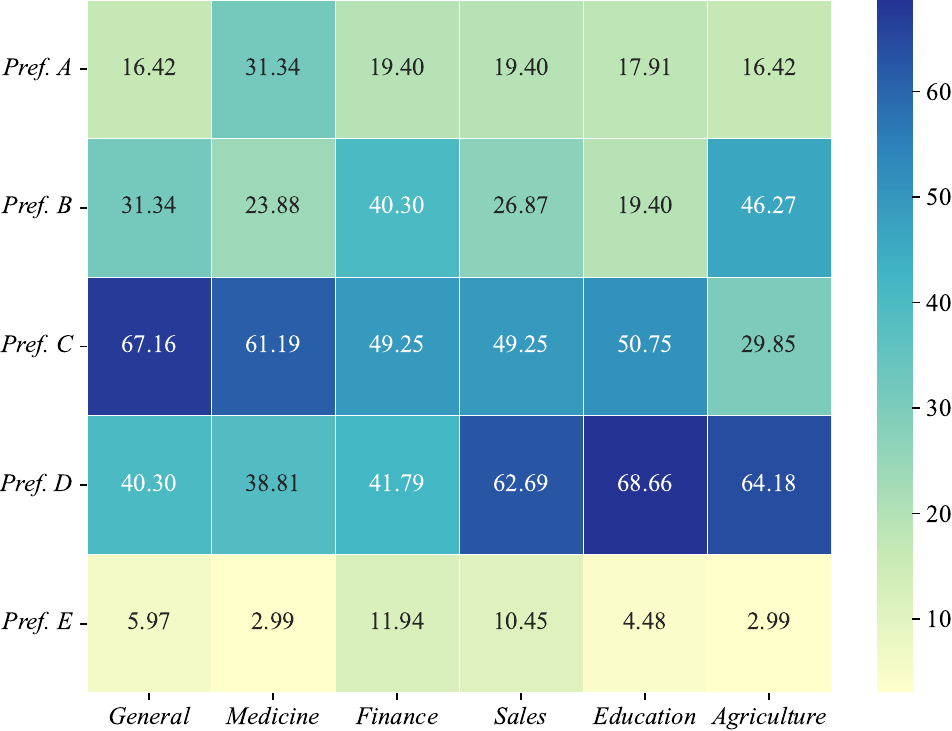}
    \caption{Heatmap for general user preference research. The more users that select a certain preference, the darker the color of the corresponding area. The values in these areas indicate the percentage of users who chose that preference for the corresponding scenario.}
    \label{fig:user_preference_analysis}
\end{figure}

In the questionnaire, we collected information such as age, gender, industry, and nationality to ensure diverse and comprehensive samples. The content of the questionnaire and specific user distribution are shown in Appendix \ref{appendix:user-research}. The distribution of the proportion of users choosing different preferences is shown in Figure \ref{fig:user_preference_analysis}.

In addition to the general scenario, we also set up scenarios for different fields. The specific questionnaire is available at Appendix \ref{appendix:questionnaire}. It is worth noting that most users are favoring the cautious preference \textit{C} and the admirer preference \textit{D}. Preference \textit{C} corresponds to the solution with the highest success possibility. This user preference is consistent with our goal of expecting more robust CEs on variable ML systems. Preference \textit{D} to some extent inspired our approach of generating CEs based on prototype samples.

\section{Datasets}
\label{appendix:dataset}
The introduction and processing details of the datasets are as follows:
\begin{itemize}
    \item \textit{Adult Income}. The dataset contains information on population, education, etc., based on the 1994 Census database and is available from the UCI Machine Learning Repository \cite{Dua:2019}. In this paper, a pre-processed version from \cite{mothilal2020explaining} was chosen to filter eight of the features. The task of the classification model is to classify whether the individual income of each sample exceeds 50,000 dollars.
    \item \textit{German Credit}. The information in this dataset is obtained from banks in relation to personal loans, such as the number of credit cards currently held by a particular bank, the duration of current employment, and other information on a total of 20 characteristics. We use the version obtained directly from the UCI database without processing. The task of the classification model is to determine whether a user is a credit risk or not by determining their credit type based on their attributes.
    \item \textit{Titanic}. This dataset is derived from the Titanic incident, in which the gender, age, ticket number, class of ticket, and other attributes of some of the passengers were collected. The dataset contains information about 891 passengers in total, and we removed the instances containing null values. The task of the classification model is to determine whether a passenger will eventually survive.
    \item \textit{Water Quality}. This dataset contains water quality measurements and assessments related to potability. Each sample contains nine attributes of one water sample, such as pH, hardness, etc. The dataset contains data for a total of 3276 water instances, and we removed some of the instances that contained null values. The task of the classification model is to discriminate whether a water sample is potable or not.
    \item \textit{Phoneme}. The dataset collects five different attributes from 1809 isolated syllables to characterize each vowel. The dataset contains a total of 5404 instances. The task of the classification model is to distinguish between nasal and oral vowels.
\end{itemize}

\section{US-Agents}
\label{appendix:us-agents}
In this section, we provide details on the implementation of user simulation agent selection experiments tailored to general user preferences.
\subsection{Profiles and Other Settings}
This is an example of a US-Agent.
\begin{tcolorbox}[colback=yellow!10!white, colframe=brown!70!olive, title=Prompts for Simulated User Experiments, breakable, fontupper=\scriptsize]
\#profile\#
Name: Alex Chen
Birthday: June 15, 1995
First person: I
Birthplace: Beijing, China
Place of residence: Shanghai, China
Occupation: Software Engineer
MBTI: ENFJ

\#Characteristics of speaking style\#
Enthusiastic and encouraging language: Uses positive and uplifting phrases to motivate and inspire others.
Focus on teamwork and collaboration: Often mentions group efforts and the importance of working together.
Balances technical jargon with accessible explanations: Able to discuss complex software concepts while making them understandable for non-experts.

\#End-of-sentence pattern\#
Frequent use of "right?" and "don't you think?" to seek agreement and engagement.
Use of exclamation marks to convey excitement and enthusiasm.
 Occasional use of ellipses (...) to pause for emphasis or to invite a response.

\#Symbol usage pattern\#
Frequent use of parentheses ( ) adds additional context or clarification.
Use bullet points or numbered lists to organize thoughts clearly.
Appropriate use of emojis to convey emotions \twemoji{smiling face} \twemoji{rocket}.

\#Example of statement\#
"Hey team, I think we're making fantastic progress on this project! Let's keep up the great work and tackle those bugs together, right?"
"I've been diving into this new framework, and it's really intuitive. I think it could streamline our workflow significantly. What do you guys think?"
"Remember, communication is key! If anyone needs help or has ideas to share, don't hesitate to speak up. We're all in this together!"
"Just had a brilliant brainstorming session with the UI team. Their insights on user experience are spot on. Collaboration really does make us stronger!"
"I believe in each one of you. Your contributions are invaluable, and together, we can achieve amazing things. Let's keep pushing forward!"
"Software development isn't just about coding; it's about solving real-world problems. That's what makes our work so rewarding, don't you think?"
"Let's not forget to celebrate our small victories. Every milestone we hit brings us closer to our goal. Keep up the fantastic work, team!"
"I'm always here to help if anyone gets stuck. We're a team, and no one gets left behind. Just reach out if you need a hand!"
"Seeing our project come to life is such a thrill. It's a testament to our hard work and dedication. Let's keep this momentum going!"
"Your ideas are brilliant, and I think they could really enhance our project. Let's discuss how we can integrate them seamlessly. Exciting times ahead!"

\#Examples of things you might not say\#
"I prefer working alone; team projects are too distracting."
"Technical details are irrelevant; just focus on the big picture."
"I don't need anyone's input; I can handle this on my own."
"Let's skip the team meeting; it's a waste of time."
"I don't care about the team's feelings; we just need to get the job done."
"Your ideas are okay, but mine are better."
"Communication is overrated; just do your part and stay out of the way."
"I don't have time to help others; I need to focus on my own tasks."
"Team morale isn't important; results are all that matter."
"Let's ignore feedback; we know what we're doing."
\end{tcolorbox}
\subsection{Prompts for Simulated User Experiments}
To guide US-Agents in performing the preference selection task, we designed the following prompt:
\begin{tcolorbox}[colback=yellow!10!white, colframe=brown!70!olive, title=Prompts for Simulated User Experiments, breakable, fontupper=\scriptsize]
Now, you have failed at a task, and two methods provide counterfactual explanations.

Counterfactual explanations are solutions that can guide you to succeed in the task. Here are some of their properties for reference:

Method A: \{tcol\_properties\}
Method B: \{baseline\_properties\}

In these properties:

- **Proximity** indicates the feature distance between the counterfactual explanation and your current situation. It can be understood that the lower the proximity value, the easier it is to implement the solution.  

- **Centrality** indicates the feature distance between the counterfactual explanation and the cluster center in the feature space. It can be understood that the lower the centrality value, the more successful cases similar to this solution exist.  

- **Sparsity** indicates the number of differing features between the counterfactual explanation and your current situation. It can be understood that the lower the sparsity value, the fewer aspects need to be changed.  

- **Validity** indicates the probability that the counterfactual explanation is classified as the desired category by the validation model. It can be understood that, assuming the validation model remains unchanged, the higher the effectiveness, the higher the success rate.  

- **Data Fidelity** indicates the robustness of the counterfactual explanation under changes in the validation model. It can be understood that the higher the data fidelity, the better the counterfactual explanation can handle model updates and changes, leading to a higher success rate.

Your preference is: \{preference\_dict[preference]\}

Please carefully consider the properties of these two methods and choose one based on your preferences.

Only choose one of the following choices: ['A', 'B'], response as JSON format: {{"method": "A" or "B"}}

\end{tcolorbox}

\end{document}